%% file: main.tex
\documentclass{article}

\PassOptionsToPackage{sort,compress}{natbib}
\usepackage{natbib}

\usepackage[utf8]{inputenc} 
\usepackage[T1]{fontenc}    
\usepackage{url}            
\usepackage{booktabs}       
\usepackage{amsfonts}       
\usepackage{amsmath}
\usepackage{nicefrac}       
\usepackage{microtype}      
\usepackage{xcolor}         
\usepackage{macros}
\usepackage{subcaption}
\usepackage{booktabs}
\usepackage{caption}
\usepackage{amsthm}
\usepackage{float}
\usepackage{wrapfig}
\usepackage{libertine}

\definecolor{mydarkblue}{rgb}{0,0.08,0.45}
\usepackage[hidelinks,
colorlinks=true,
linkcolor=mydarkblue,
citecolor=mydarkblue,
filecolor=mydarkblue,
urlcolor=mydarkblue,
backref=section]{hyperref}       
\usepackage{cleveref}

\title{On Kernel Regression with Data-Dependent Kernels}
\author{
James B. Simon \\
\\
UC Berkeley \& Generally Intelligent \\
\texttt{\sf james.simon@berkeley.edu} \\
}
\date{}

\begin{document}

\maketitle


\centerline{\fontsize{13}{15}\selectfont \textbf{Abstract}}

The primary hyperparameter in kernel regression (KR) is the choice of kernel.
In most theoretical studies of KR, one assumes the kernel is fixed before seeing the training data.
Under this assumption, it is known that the optimal kernel is equal to the prior covariance of the target function.
In this note, we consider KR in which the kernel may be updated \textit{after} seeing the training data.
We point out that an analogous choice of kernel using the \textit{posterior} of the target function is optimal in this setting.
Connections to the view of deep neural networks as data-dependent kernel learners are discussed.



\input{sections/intro}

\input{sections/preliminaries}

\input{sections/main_section}

\input{sections/conclusions}

\section*{Acknowledgements}
The author thanks Berfin \c{S}im\c{s}ek, Preetum Nakkiran, Daniel Beaglehole, Song Mei, and Mike DeWeese for useful discussions and the 2022 Les Houches Summer School on Statistical Physics and Machine Learning for fostering discussion that led to this note.

\bibliographystyle{plainnat}
\bibliography{ml_refs}

\appendix
\onecolumn

\input{appendices/prior_proofs}
\input{appendices/posterior_proofs}

\input{appendices/experimental_details}


\end{document}

%% file: sections/intro.tex
\section{Introduction}
\label{sec:intro}

Kernel regression (KR) is a widespread nonparameteric learning algorithm which has seen a surge of attention due to equivalences to wide neural networks \citep{lee:2018-nngp, matthews:2018-nngp,  jacot:2018, lee:2019-ntk}.
In the KR setting standard in theoretical study, the kernel is assumed to be a static hyperparameter, fixed before seeing training data, and typical theoretical results describe e.g. the generalization of KR with a certain kernel on a certain family of target functions.
In this note, we instead consider KR in which the kernel can be chosen adaptively depending on the training data, identifying a simple choice for an optimal such data-dependent kernel.
Furthermore, KR with this kernel in fact yields the mean of the posterior, i.e. the \textit{best possible predictor} given observed data.

As motivation for this adaptive setting, note that a variety of KR variants, such as those called ``multiple kernel learning," perform this kernel adaptation explicitly \citep{gonen:2011-multiple-kernel-learning, cristianini:2001-kernel-target-alignment, cortes:2012-learning-kernels-based-on-alignment, sinha:2016-learning-rf-kernels}.
Furthermore, in practice, the kernel will typically be chosen via cross-validation, and thus it is implicitly adapted to the training data.
In addition, recent work suggests that practical neural networks, rather than representing fixe
d kernels during training, can instead be roughly decomposed into the two stages of (1) learning a data-dependent kernel and (2) performing KR with said kernel, which matches our setting \citep{fort:2020-deep-learning-versus-kernel-learning, atanasov:2021-silent-alignment, long:2021-after-kernel, vyas:2022-limitations-of-the-ntk-for-generalization}.
The optimal posterior kernel we identify is not typically feasible to compute in practice (as it requires knowledge of the posterior distribution over targets), but we might nonetheless hope that, by studying an \textit{optimal} choice of adaptive kernel, we can identify desirable structures and properties to look for in practical adaptive KR methods.

We briefly give preliminaries in Section \ref{sec:preliminaries}.
We state our main result and discuss connections to neural networks in Section \ref{sec:main_section} and conclude in Section \ref{sec:conclusions}.

%% file: sections/preliminaries.tex
\section{Preliminaries}
\label{sec:preliminaries}

Consider the standard supervised regression task of learning an unknown scalar target function from training data.
Suppose we are provided a training set of $n$ inputs $\X = \{x_i\}_{i=1}^n$ drawn i.i.d. from a measure $\mu_x$ over $\R^d$ and corresponding targets $\Y = \{f(x_i)\}_{i=1}^n$, where $f$ is the target function.
For brevity, we will denote the full dataset as $\D \equiv (\X,\Y)$.
Suppose further that the target function itself is sampled from a measure $\mu_f$ over an appropriate function space\footnote{We shall assume the target function $f$ is deterministic in the sense that, once $f$ is fixed, then $f(x)$ for a given $x$ is the same each time it is queried. However, for an appropriate prior $\mu_f$, the function $f$ can still be effectively ``noisy" in the sense of containing a component with zero correlation length, with the noise ``baked into" $f$ instead of sampled anew with each query. This distinction only matters in the unlikely case in which one sees the same input $x$ twice, and so we neglect it and do not trouble ourselves with the question of noise.}.

We shall consider using KR to extend this function to unknown test data.
KR prescribes the predicted function
\begin{equation} \label{eqn:kr}
    \hat{f}(x) = \kb_{x\X} \Kb_{\X\X}^{-1} \Y,
\end{equation}
where, using our chosen kernel function $K(\cdot,\cdot)$, we have constructed the row-vector $[\kb_{x\X}]_i = K(x,x_i)$ and the data-data kernel matrix $[\Kb_{\X\X}]_{ij} = K(x_i,x_j)$, we interpret $\Y$ as a column vector, and the matrix inverse is the Moore-Penrose pseudoinverse.
As our objective, we shall aim to minimize the expected squared risk
\begin{equation}
\e \equiv \E_{f \sim \mu_f} \E_{x \sim \mu_x} \left[ \left( \hat{f}(x) - f(x) \right)^2 \right].
\end{equation}

%% file: sections/main_section.tex
\section{Prior and Posterior Kernels}
\label{sec:main_section}

We will contrast two paradigms for choosing the kernel $K$:
\begin{itemize}
    \item \textbf{Setting 1:} $K$ is chosen \textit{a priori}, before seeing $\D$.
    This is the setting typically studied.
    \item \textbf{Setting 2:} $K$ can be modified after seeing $\D$.
    This setting permits a data-informed choice of kernel and is the focus of this note.
\end{itemize}
In both settings, we shall study the optimal choice of kernel as a function of (and assuming knowledge of) $\mu_x$ and $\mu_f$.

\subsection{The Optimal Prior Kernel}

The optimal kernel in Setting 1 is easily identified using a simple Bayesian calculation (see e.g. Appendix B.7 of \citet{jacot:2020-KARE}).
The kernel minimizing expected test risk in Setting 1 is the \textit{prior kernel} given by
\begin{equation} \label{eqn:K_prior}
    \Kprior(x,x') \equiv \E_{f \sim \mu_f}[f(x)f(x')].
\end{equation}
If $\mu_f$ is \textit{centered} (in the sense that $\E_{f \sim \mu_f}[f(x)] = 0$ for all $x$), then this kernel is simply the prior covariance of $f$.

This kernel is optimal in the following sense:
\begin{proposition}[Optimal prior kernel]
\label{prop:prior}
In Setting 1, KR with kernel $\Kprior$ achieves the minimum expected test risk of any predictor with \emph{arbitrary} dependence on $\X$ and \emph{linear} dependence on $\Y$.
\end{proposition}
We provide a proof of Proposition \ref{prop:prior} in Appendix \ref{app:prior}.

\textbf{Remark:} if $\mu_f$ is not centered, then one can trivially center it by subtracting off the mean, then predicting $f(x) - \E_f[f(x)]$ and adding back the mean post hoc.
KR with this offset and kernel equal to the prior covariance of $f$ is then the optimal predictor with \textit{affine} dependence on $\Y$.
We formally state and prove this extension in Appendix \ref{app:prior}.

\textbf{Remark.} If the prior $\mu_f$ is a centered Gaussian process (GP) with kernel $K$, then not only is $\Kprior = K$ the optimal prior kernel, but KR with kernel $K$ yields the mean of the posterior, the optimal predictor given the data.
In this case, we can stop our search for a better kernel here.
However, if the priors are \textit{not} a GP, we can hope to gain something by intelligently adapting the kernel to the data.

\subsection{The Optimal Posterior Kernel}

We now turn to Setting 2, in which the kernel is chosen after observation of the dataset $\D$, including the training labels $\Y$.
We will identify an analogous optimal kernel in this setting.
This kernel will not yield a practically useful algorithm, but we might nonetheless hope to study it to identify signatures of good adaptive kernels.
We present the kernel here and discuss what we might learn from it in the following subsection.

We might guess, by analogy to Equation \ref{eqn:K_prior}, that a promising kernel in this setting is what one might call the \textit{posterior kernel}
\begin{equation} \label{eqn:K_posterior}
    \Kpost(x,x) \equiv \E_{f \sim \mu_f | \D}[f(x)f(x')].
\end{equation}
In Equation \ref{eqn:K_posterior}, we have replaced the expectation over the target prior $\mu_f$ with an expectation over the posterior, with $\E_{f \sim \mu_f | \D}[\cdot] \equiv \E_{(f \sim \mu_f | f(\X)=\Y)}[\cdot]$ denoting an average over targets $f$ conditioned on the fact that $f(\X) = \Y$.

Remarkably, this guess is optimal in an even stronger sense than the prior kernel in the classic setting.
This is encapsulated in the following proposition:

\begin{proposition}[Optimal posterior kernel]
\label{prop:post}
In Setting 2, KR with kernel $\Kpost$ yields $\hat{f}(x) = \E_{f \sim \mu_f | D}[f(x)]$, the mean of the posterior of $f$, and thus achieves the minimum expected test risk of any predictor with arbitrary dependence on $\X$ \emph{and} $\Y$.
\end{proposition}

The proof of Proposition \ref{prop:post} is quite simple, and we provide it here.

\textbf{Proof of Proposition \ref{prop:post}.} 
First, we observe that, for $x_i \in \X$, we have that $f(x_i)$ is deterministically equal to $\Y_i$ when $f$ is sampled from the posterior measure conditioned on $\D$.
This implies that, for all $x_i, x_j \in \X$,
\begin{align}
&\Kpost(x_i,x_j) = \E_{f \sim \mu_f | \D}[f(x_i) f(x_j)] = \Y_i \Y_j, \label{eqn:YY} \\
&\Kpost(x,x_i) = \E_{f \sim \mu_f | \D}[f(x) f(x_i)] = \Y_i \, \E_{f \sim \mu_f | \D}[f(x)]. \label{eqn:YEf}
\end{align}
Equations \ref{eqn:YY} and \ref{eqn:YEf} in turn imply that kernel $\Kpost$ gives kernel matrices
\begin{align}
\Kbpost_{\X\X} &= \Y \Y^\top, \label{eqn:gram_post} \\
\kbpost_{x\X} &= \Y^\top \, \E_{f \sim \mu_f | \D}[f(x)]. \label{eqn:k_post}
\end{align}

Using Equation \ref{eqn:kr} and taking note of the matrix pseudoinverse, we then find that
\begin{equation}
    \hat{f}(x) = \E_{f \sim \mu_f | \D}[f(x)] \cdot \Y^\top (\Y \Y^\top)^{-1} \Y = \E_{f \sim \mu_f | \D}[f(x)],
\end{equation}
which is the mean $f(x)$ with respect to the posterior over $f$.
It is a property of squared risk that this predictor achieves the minimum expected squared risk over all predictors depending arbitrarily on the observed data.
\pushQED{\qed}
\popQED

\textbf{Remark:} The posterior kernel in fact yields the optimal predictor in even more general KR settings than in our framing.
For example, even if the data measure $\mu_x$ is itself a random variable correlated with $f$, or even if the train inputs $\X$ are drawn from a different distribution than the test point $x$, the posterior kernel (with appropriate conditionalization) yields the optimal predictor.
We formulate and prove these generalizations in Appendix \ref{app:posterior}.
We note that, unlike in the posterior case (Setting 2), statistical dependence between $\X$ and $f$ breaks Proposition \ref{prop:prior} for the prior case (Setting 1): KR with the prior kernel $\Kprior$ is no longer the best predictor with linear dependence on $\Y$ and arbitrary dependence on $\X$ because we have neglected clues $\X$ might give as to $f$.

The posterior kernel $\Kpost$ is an unusual kernel and merits some discussion.
As a first observation, note that, while it can have highly nontrivial structure \textit{off} the training set, it is trivial (i.e. rank-one) \textit{on} the training set, as evidenced by Equation \ref{eqn:YY}.
This perfect ``memorization" of $\Y$ is what leads to the optimal posterior predictor: as seen in the proof of Proposition \ref{prop:post}, KR with kernel $\Kpost$ essentially acts to project $\Y$ directly onto the mean of the posterior.

It is worth noting that, while KR with $\Kpost$ is optimal, $\Kpost$ is not the \textit{only} kernel for which this is true.
For example, modifying $\Kpost$ on $\X$ such that $\Kbpost_{\X\X}$ gains additional eigenvectors orthogonal to $\Y$ has no effect on the predicted function $\hat{f}$.
While not unique, however, $\Kpost$ is perhaps in some sense the simplest choice.

\subsection{Discussion}
\label{subsec:discussion}

What can we learn from Proposition \ref{prop:post}?
Astute readers might observe that, since we assumed (and leveraged) full knowledge of task priors, we might have dispensed with kernels entirely and just directly used the mean of the posterior.
If there is interest to this result, then, it is perhaps that it suggests some properties we might expect to see in good data-dependent kernels in the wild, such as in neural networks.
In fact, it is easy to identify two such features --- kernel alignment with targets and low rank on the data --- which neural networks' post-training kernels have indeed been found to exhibit.

\textbf{Kernel alignment with $\Y$.}
It is well-accepted that kernel ``alignment" with the target function is correlated with good generalization \citep{cristianini:2001-kernel-target-alignment, bordelon:2020-learning-curves, jacot:2020-KARE}.
The optimal posterior kernel $\Kpost$ is in fact \textit{maximally aligned} with the training labels $\Y$ in the sense that the alignment metric
\begin{equation} \label{eqn:a_YK}
    a_{\Y,K} \equiv \frac{\Y^\top \Kb \Y}
    {|\!| \Y |\!|^2 \ |\!| \Kb |\!|_F}
\end{equation}
is maximized, and also in the sense that metric
\begin{equation}
    \tilde{a}_{\Y,K} \equiv \frac{\Y^\top \Kb^{-1} \Y}
    {|\!| \Y |\!|^2 \ |\!| \Kb^{-1} |\!|_F}
\end{equation}
is minimized\footnote{To motivate the quantity $\tilde{a}_{\Y,K}$, note that the RKHS norm of $\hat{f}$ with respect to kernel $K$ is $|\!|\hat{f}|\!|_{K} = \Y^\top \Kb^{-1} \Y$.}.
\citet{baratin:2021-ntk-alignment} and \citet{atanasov:2021-silent-alignment} measured the alignment $a_{\Y,\Kpost}$ throughout training of the training labels and the data-data NTK of MLPs, VGG convolutional networks, and ResNets on MNIST and CIFAR-10, finding in all cases that alignment dramatically increases, especially early in training.


\begin{figure}[h]
    \centering
    \includegraphics[width=8cm]{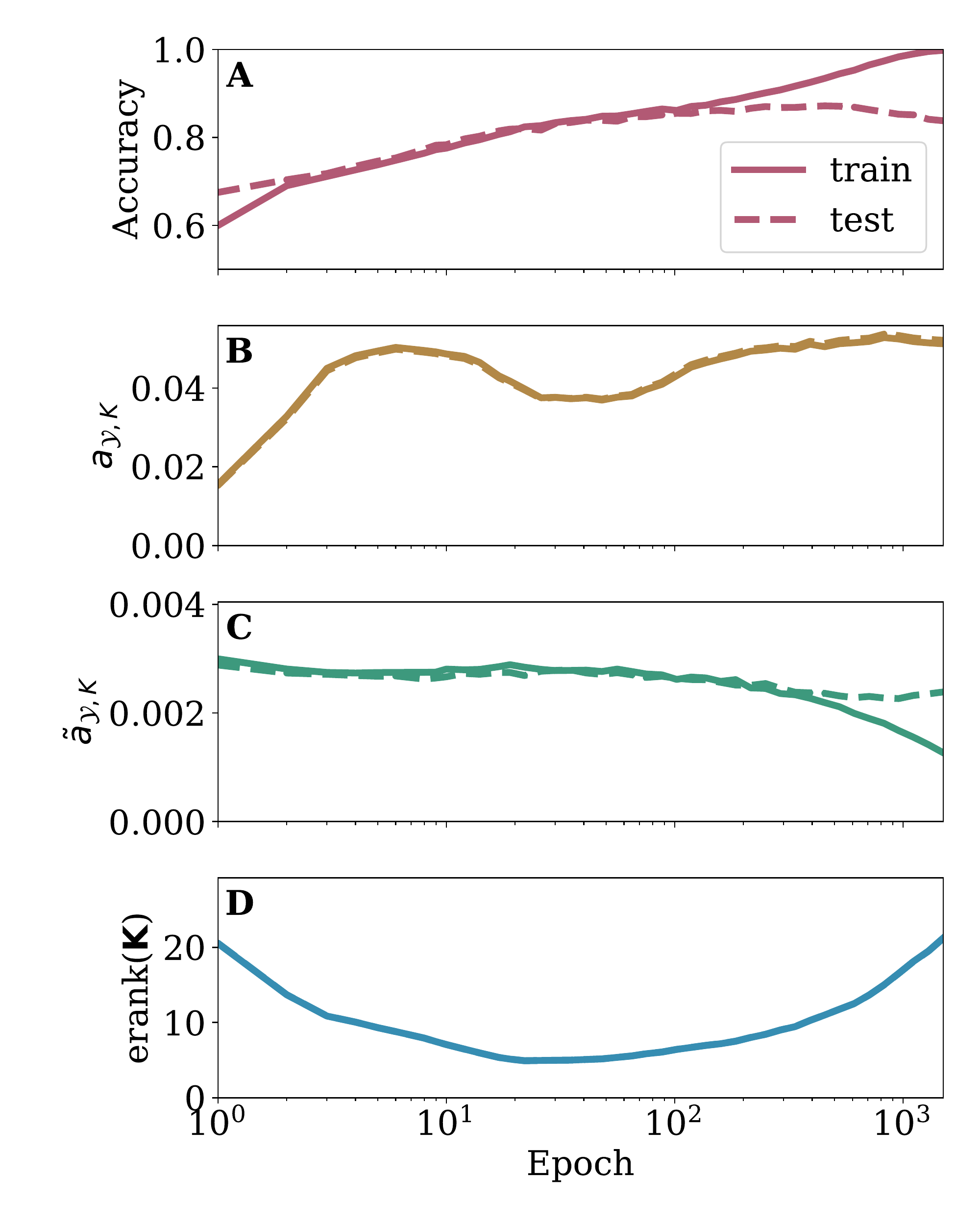}
    \caption{
    Accuracy and NTK alignment metrics of a basic CNN trained on a binarized version of CIFAR-10 with 8k train and 8k test samples.
    \textbf{(A)} Train and test accuracy.
    \textbf{(B)} Train and test empirical kernel alignment $a_{\Y,K}$.
    Higher $a_{\Y,K}$ means greater alignment.
    \textbf{(C)} Train and test normalized RKHS norm $\tilde{a}_{\Y,K}$.
    Lower $a_{\Y,K}$ means greater alignment.
    \textbf{(D)} Effective rank of the empirical kernel.
    }
    \label{fig:alignment_plots_main}
\end{figure}

\textbf{Low-rank data-data kernel matrix.}
The optimal posterior kernel $\Kpost$ is such that $\Kbpost_{\X\X}$ is significantly reduced in rank.
A similar phenomenon can occur with neural network kernels early in training.
\citet{baratin:2021-ntk-alignment} examined the eigenvalues throughout training of the data-data NTK matrix for a VGG convolutional network on CIFAR-10 and found that, as learning proceeds, the top few eigenvalues grow disproportionately and the effective rank of the kernel matrix sharply drops.
Though these dynamics are poorly understood, this decrease in rank makes sense in light of the fact that it enables an increase in alignment if the emergent top eigenmodes are thus aligned.
The optimal posterior kernel provides a clear example where this is the case, reinforcing the idea that perhaps we ought to expect good posterior kernels to be spikier than their prior counterparts.

\textbf{CNN experiment.}
For completeness of narrative, an experiment is included illustrating both the increase in alignment and initial decrease in effective rank using a simple CNN on a binarized size-8k subset of CIFAR-10 with MSE loss\footnote{We binarize CIFAR-10 into two superclasses, \texttt{airplane-automobile-horse-ship-truck} and \texttt{bird-cat-deer-dog-frog} (i.e. things one can safely ride and things one had better not).}.
We evaluate performance and kernel metrics on both the training set and a held-out test set of the same size.
Following \citet{baratin:2021-ntk-alignment}, if the eigenvalues of a matrix $\Kb$ are $\{\lambda_i\}_i$ and we let $\mu_i \equiv \lambda_i / (\sum_i \lambda_i)$, we define the effective rank to be
\begin{equation}
    \text{erank}(\Kb) \equiv \exp \left\{ -\sum_i \mu_i \log \mu_i\right \}.
\end{equation}
The results are in Figure \ref{fig:alignment_plots_main}, and experimental details and a notebook to reproduce the figure are provided in Appendix \ref{app:exp}.
We see an increase in $a_{\Y,K}$ and decrease in $\tilde{a}_{\Y,K}$, implying growing kernel alignment, with alignment greater on the train than on the test set.
Additionally, the effective kernel rank falls quickly and is small when test accuracy approaches its maximum.
Somewhat surprisingly, we find that these trends are not monotonic; $a_{\Y,K}$ exhibits a short period of decrease, and the effective rank \textit{grows} during overfitting towards the end of training.
This late-time increase in effective rank was also seen in experiments by \citet{baratin:2021-ntk-alignment} using noisy data, and in both their and our experiments, this increase coincides with the saturation of test accuracy and overfitting of the training data, suggesting that it may be undesirable.

It should be noted that the effects we see are much weaker than those of \citet{baratin:2021-ntk-alignment} and \citet{atanasov:2021-silent-alignment}, who each observed alignment coefficients $a_{\Y,K}$ higher than $0.2$, a difference that can likely be attributed to our simpler setting.

%% file: sections/conclusions.tex
\section{Conclusions}
\label{sec:conclusions}

We set out to understand the optimal choice of kernel for KR, if we permit modification of the kernel after seeing the training data.
We identified such an optimal choice in the kernel $\Kpost$, given by a natural formula analogous to that of the optimal \textit{a priori} kernel, and noted that it exhibits alignment properties loosely similar to those seen in trained NTKs, which suggests that they might be general phenomena to expect when choosing good \textit{a posteriori} kernels.
There are, of course, plenty of caveats to drawing too tight a correspondence: to name one, networks' NTKs increase in \textit{trace} as well as alignment, while this isn't necessarily true of the posterior kernel.
Nonetheless, it is attractive to think that a trained NTK might somehow be thought of as ``conditioned" on the data, so perhaps this correspondence is worth keeping in mind.

%% file: appendices/prior_proofs.tex
\section{Prior Kernel Proofs}
\label{app:prior}

In this Appendix, we first give a proof of Proposition \ref{prop:prior} that $\Kprior$ is the optimal prior kernel, then state and prove an analogous proposition applicable to an affine version of KR.

Before we begin, we introduce some helpful notation.
Recall that the prior kernel $\Kprior(x_1,x_2) \equiv \E_{f \sim \mu_f}[f(x_1)f(x_2)].$
We shall define the mean function $\bar{f}(x) \equiv \E_{f \sim \mu_f}[f(x)]$ and the prior \textit{covariance} kernel
$\Kcov(x_1,x_2) \equiv \E_{f \sim \mu_f}[(f(x_1) - \bar{f}(x_1))(f(x_2) - \bar{f}(x_2))].$
To construct matrices and vectors from these functions, we will use a notation different from that of the main text, writing e.g. $\Kprior(x,\X)$ for the row vector with $[\Kprior(x,\X)]_i = \Kprior(x,x_i)$.
We will similarly write $\Kprior(\X,\X)$ for the data-data kernel matrix and $\bar{f}(\X)$ for the column vector of function evaluations.

\textbf{Proof of Proposition \ref{prop:prior}.}
Proposition \ref{prop:prior} states that KR with kernel $\Kprior$ achieves minimal test risk over all predictors with arbitrary dependence on $\X$ but linear dependence on $\Y$.
To show this, we shall write down the most general such predictor, enforce an optimality condition, and arrive at KR with this kernel.

The most general linear predictor is
\begin{equation}
    \hat{f}(x) = \ab^\top f(\X),
\end{equation}
where $\ab$ is a $n$-element vector with arbitrary dependence on $x$ and $\X$.
The expected squared error on point $x$ is
\begin{equation}
    \e(x) = \E_{f \sim \mu_f}[(f(x) - \ab^\top f(\X))^2].
\end{equation}
Differentiating with respect to $\ab$, we find that, at optimality,
\begin{align}
    \nabla_{\ab} \e(x)
    &= \E_{f \sim \mu_f}[2 \ab^\top f(\X) f^\top(\X) - 2 f(x) f^\top(\X)] \\
    &= 2 \ab^\top \Kprior(\X,\X) - 2 \Kprior(x,\X) \\
    &= 0.
\end{align}
Solving for $\ab$, we find that
\begin{equation}
    \ab = \Kprior(\X,\X)^{-1} \Kprior(\X,x) \ \ \ \Rightarrow \ \ \ \hat{f}(x) = \Kprior(x,\X) \Kprior(\X,\X)^{-1} f(\X),
\end{equation}
which is precisely KR with kernel $\Kprior$. \pushQED{\qed}\popQED

If $\hat{f}(x) \neq 0$, then intuitively we might expect to benefit from predicting $f(x) - \hat{f}(x)$ with KR instead of predicting $f(x)$ directly.
It turns out that a predictor of this fashion with kernel $\Kcov$ is in fact the optimal predictor affine in the training targets (and thus the benefit of subtracting off the mean is nonnegative).
This is formalized in the following proposition:

\begin{proposition}
\label{prop:affine_prior}
The ``affine KR" predictor with kernel $\Kcov$, defined as
\begin{equation} \label{eqn:affine_kr}
\hat{f}(x) = \Kcov(x,\X) \Kcov(\X,\X)^{-1} \left( f(\X) - \bar{f}(\X) \right) + \bar{f}(\X),
\end{equation}
achieves the minimum expected squared risk of any predictor with arbitrary dependence on $\X$ and affine dependence on $f(\X)$.
\end{proposition}

\textbf{Proof of Proposition \ref{prop:affine_prior}.}
As in the previous proof, we first write down the most general ``affine predictor"
\begin{equation}
    \hat{f}(x) = \ab^\top f(\X) + b,
\end{equation}
where both the vector $\ab$ and scalar $b$ can again have arbitrary dependence on $x$ and $\X$.
Anticipating our next step, we will first reparameterize this formula as
\begin{equation}
    \hat{f}(x) - \bar{f}(x) = (\ab')^\top (f(\X) - \bar{f}(\X)) + b.
\end{equation}
We then have expected squared error
\begin{equation}
    \e(x) = \E_{f \sim \mu_f}\left[\left(f(x) - \bar{f}(x) - (\ab')^\top (f(\X) - \bar{f}(\X)) - b\right)^2\right].
\end{equation}
Enforcing optimality with respect to $\ab'$ and $b$ yields that
\begin{align}
    \nabla_{\ab'} \e(x) &= 2 \Kcov(\X,\X) \ab' - 2 \Kcov(x,\X) = 0, \\
    \nabla_{b} \e(x) &= 2b = 0.
\end{align}
Solving these equations tells us that $\ab' = \Kcov(\X,\X)^{-1} \Kcov(\X,x)$ and $b = 0$, yielding the optimal affine KR predictor of Equation \ref{eqn:affine_kr}. \pushQED{\qed} \popQED

%% file: appendices/posterior_proofs.tex
\section{Posterior Kernel Proofs}
\label{app:posterior}

In this Appendix, we prove a generalization of Proposition \ref{prop:post} on the optimality of the prior kernel.
In the main text, we considered settings in which the train and test data are drawn from a fixed, known measure $\mu_x$.
Here we consider a more general setting in which the train and test distributions can be different from each other and are themselves random variables possibly correlated with the target function.

Formally, instead of drawing train and test data i.i.d. from a known measure $\mu_x$, they are sampled from unknown measures $\mu_x^{\text{tr}}$ and $\mu_x^{\text{te}}$.
These measures, and the target function, are together sampled from a ``meta-measure" $\mu^*$ as $(f, \mu_x^{\text{tr}}, \mu_x^{\text{te}}) \sim \mu^*$.

Since $\mu_x^{\text{tr}}$ and $\mu_x^{\text{te}}$ can be correlated with $f$, the training set $\X$ and test point $x$ can offer clues as to the identity of $f$, and we must note this in our marginalization.
The posterior mean (and optimal predictor) is now
\begin{equation} \label{eqn:f_opt_gen}
    \hat{f}_{\text{opt}}(x) = \E_{(f, \mu_x^{\text{tr}}, \mu_x^{\text{te}}) \sim \mu^* | \D, x} [ f(x) ],
\end{equation}
where the expectation is over tasks conditioned on observing the provided $\X$, $\Y$, and $x$.
We accordingly redefine the optimal posterior kernel as
\begin{equation}
    \Kpostgen(x,x') = \E_{(f, \mu_x^{\text{tr}}, \mu_x^{\text{te}}) \sim \mu^* | \D, x} [ f(x) f(x') ].
\end{equation}

The following version Proposition \ref{prop:post} then holds:
\begin{proposition} \label{prop:post_gen}
    In Setting 2, KR with kernel $\Kpostgen$ yields the optimal predictor of Equation \ref{eqn:f_opt_gen}.
\end{proposition}

\textbf{Proof of Proposition \ref{prop:post_gen}.}
The proof is virtually identical to that of Proposition \ref{prop:post}.
The data-data kernel matrix is again $\Kpostgen(\X,\X) = \Y \Y^\top$ and the projection row-vector is $\Kpostgen(x,\X) = \hat{f}_{\text{opt}}(x) \Y^\top$, yielding $\hat{f}(x) = \hat{f}_{\text{opt}}(x)$. \pushQED{\qed} \popQED

This result can in fact be extended to an even more general setting in which, instead of being drawn i.i.d. from a training measure $\mu_x^{\text{tr}}$, the training points are not drawn i.i.d., and instead the inputs $\X$ are drawn as a set from a distribution $\mu_x^{\text{tr},n}$.

%% file: appendices/experimental_details.tex
\section{Experimental Details}
\label{app:exp}

The experiment of figure \ref{fig:alignment_plots_main} uses a simple CNN with three convolutional layers with 32 hidden channels and $5 \times 5$ filters and a linear readout layer.
The training data consisted of 8k samples from CIFAR-10 binarized into superclasses \texttt{airplane-automobile-horse-ship-truck} and \texttt{bird-cat-deer-dog-frog}, with scalar $\pm 1$ targets.
The test data similarly consisted of 8k samples.
The network was trained via SGD for $1500$ epochs with batch size 128 with learning rate $2 \times 10^{-5}$.
This is a small learning rate for this network and task, which made the alignment effects easier to see.

The test-train accuracy gap on the left side of Figure \ref{fig:alignment_plots_main}A is due to the fact that train metrics are computed throughout the epoch, but test metrics are computed \textit{after} the epoch, so test accuracy is roughly ``half an epoch higher" early in training.
In computing the effective rank, we lower-bound kernel eigenvalues at $\texttt{eps} = 10^{-10}$ for numerical reasons.
In computing $\tilde{a}_{\Y,K}$, we use \texttt{torch.linalg.pinv} with relative tolerance $\texttt{rtol} = 10^{-7}$.

Finally, it bears mention that, in the generation of Figure \ref{fig:alignment_plots_main}, several hyperparameters (learning rates, dataset sizes, and class binarizations) were tried.
While some phenomena were very common, like the initial increase in alignment and initial drop of effective rank, some features of the presented curves vary: for some datasets, $a_{\Y,K}$ \textit{decreases} after a period of high value, or the bumps in the increasing curve are different.
The late-time increase in effective rank was typically smaller than reported in Figure \ref{fig:alignment_plots_main}D.
However, the decrease in $\tilde{a}_{\Y,K}$ was essentially monotonic in every trial, and usually also featured a train-test gap coinciding with the accuracy gap (compare the late-time behavior in Figures \ref{fig:alignment_plots_main}A and \ref{fig:alignment_plots_main}C).
This suggests that $\tilde{a}_{\Y,K}$ may in fact be a more informative measure of alignment than $a_{\Y,K}$ and worth inclusion in future studies of NTK alignment.

A Colab notebook reproducing Figure \ref{fig:alignment_plots_main} can be found at \url{https://colab.research.google.com/drive/1hXCPPj8Yej0-M6_SkbpuYx1JGlLqf0dM}.